\documentclass[letterpaper,journal]{IEEEtran}
\usepackage{amsmath,amsfonts}
\usepackage{algorithmic}
\usepackage{algorithm}
\usepackage{array}
\usepackage[caption=false,font=normalsize,labelfont=sf,textfont=sf]{subfig}
\usepackage{textcomp}
\usepackage{stfloats}
\usepackage{url}
\usepackage{verbatim}
\usepackage{graphicx}
\usepackage{cite}
\usepackage{booktabs}
\usepackage{multirow} 
\usepackage{bm}
\usepackage{hyperref}
\begin{document}
\pagestyle{empty}
\title{SuperGrasp: Single-View Object Grasping via Superquadric Similarity Matching, Evaluation, and Refinement}
\author{Lijingze Xiao$^{\dagger}$, Jinhong Du$^{\dagger}$, Supeng Diao, Yu Ren, Yang Cong$^{*}$
\thanks{$^{\dagger}$indicates equal contributions.}
\thanks{$^{*}$The corresponding author is Prof. Yang Cong.}
}
% \author{Anonymous Submission}



\maketitle
\thispagestyle{empty}
\begin{abstract}
% Robotic grasping from single-view observations remains a critical challenge in manipulation. Existing methods still struggle to generate stable and valid grasp poses when confronted with incomplete geometric information. To address these limitations, we propose SuperGrasp, a novel two-stage framework for single-view grasping with parallel-jaw grippers that decomposes the grasping process into initial grasp pose generation and subsequent grasp evaluation and refinement. In the first stage, we introduce a Similarity Matching Module that efficiently retrieves grasp candidates by matching the input single-view point cloud with a precomputed primitive dataset based on superquadric coefficients. In the second stage, we propose E-RNet, an end-to-end network that expands the grasp-aware region and takes the initial grasp closure region as a local anchor region, enabling more accurate and reliable evaluation and refinement of grasp candidates. To enhance generalization, we construct a primitive dataset containing 1.2k standard geometric primitives for similarity matching and collect a point cloud dataset of 100k samples from 124 objects, annotated with stable grasp labels for network training. Extensive experiments in both simulation and real-world environments demonstrate that our method achieves stable grasping performance and good generalization across novel objects and clutter scenes.
Robotic grasping from single-view observations remains a critical challenge in manipulation. However, existing methods still struggle to generate reliable grasp candidates and stably evaluate grasp feasibility under incomplete geometric information. To address these limitations, we present SuperGrasp, a new two-stage framework for single-view parallel-jaw grasping. In the first stage, we introduce a Similarity Matching Module that efficiently retrieves valid and diverse grasp candidates by matching the input single-view point cloud with a precomputed primitive dataset based on superquadric coefficients. In the second stage, we propose E-RNet, an end-to-end network that expands the grasp-aware region and takes the initial grasp closure region as a local anchor region, capturing the contextual relationship between the local region and its surrounding spatial neighborhood, thereby enabling more accurate and reliable grasp evaluation and introducing small-range local refinement to improve grasp adaptability. To enhance generalization, we construct a primitive dataset containing 1.2k standard geometric primitives for similarity matching and collect a point cloud dataset of 100k samples from 124 objects, annotated with stable grasp labels for network training. Extensive experiments in both simulation and real-world environments demonstrate that our method achieves stable grasping performance and good generalization across novel objects and clutter scenes.
\end{abstract}

\begin{IEEEkeywords}
single-view grasping, superquadric similarity matching, grasp evaluation and refinement.
\end{IEEEkeywords}

\section{Introduction}
Grasping is a fundamental capability in robotic manipulation and has shown great potential in household service and industrial production-line applications in structured environments. However, grasping objects in unstructured environments remains challenging, especially under single-view observations, due to incomplete and noisy visual input. Using such degraded geometric information to generate stable grasp poses is still difficult for existing methods.

Superquadrics\cite{barr1981superquadrics} were introduced into computer vision in the late twentieth century as a compact representation of geometric primitives. Later, with improvements to fitting algorithms by\cite{ems,Superquadrics-Revisited}, superquadrics became increasingly effective for representing object geometry directly from point clouds. This progress has inspired new solutions for robotic grasping. For example, Wu et al.\cite{wu2023learning-free} take point clouds as input, compute multiple graspable regions based on superquadric coefficients, and then sample grasp poses accordingly, achieving promising results. However, these methods are generally developed under the assumption of complete point clouds. For single-view point clouds, geometric representation still suffers from substantial difficulties in both shape and scale alignment. To alleviate this issue, Chen et al.\cite{multi} proposed a multi-level similarity matching strategy to address the difficulty of grasp candidate generation under single-view observations. Nevertheless, similarity estimation can be dominated by visible local surfaces, while the unseen global geometry is weakly constrained. As a result, the transferred grasp may fit the observed region, but still collide with the invisible part of the target object. 

With the rapid development of neural networks, data-driven grasping methods such as GPD\cite{gpd} and PointNetGPD\cite{pointnetgpd} adopt a two-stage framework. They first sample grasp candidates from point-cloud geometry and then classify these candidates using deep neural networks based on the local gripper closing region. However, due to geometric incompleteness in sparse point clouds, the sampled grasp candidates are often unreliable, and relying solely on the gripper closing region may further limit the accuracy and robustness of the subsequent evaluation module.

\begin{figure}[!t]
    \centering
    \includegraphics[width=\columnwidth]{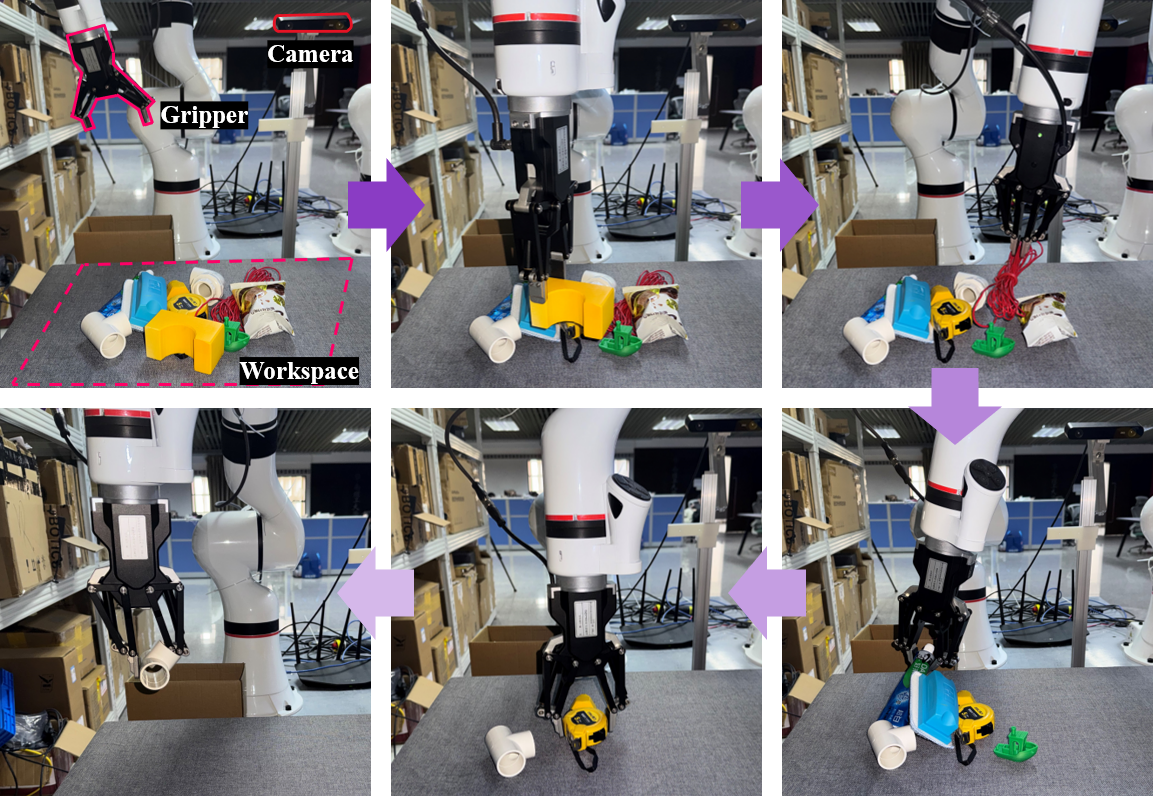}
    \caption{\hspace*{-0.6em}Overview of the task pipeline. Under dense clutter and single-view observation constraints, the tabletop decluttering task is accomplished through a two-stage pipeline consisting of grasp candidate generation, grasp evaluation, and refinement.}
    \label{fig:task-pipeline}
\end{figure}

To address these challenges, we propose SuperGrasp, a two-stage grasping pipeline similar to PointNetGPD\cite{pointnetgpd}, but with novel modules in both stages. Specifically, in the first stage, we directly estimate the superquadric coefficients from the object’s single-view point cloud to approximate its overall geometry. We then perform global similarity matching against an offline primitive database built on superquadric coefficients, where both shape and scale are considered, to enhance its geometric representation. The grasp poses of the top-$N$ most similar primitive objects are subsequently transformed into the point cloud of the target object, producing rich and effective grasp candidates.

In the second stage, to enable more reliable evaluation and refinement, we design an anchor-guided evaluation-refinement network based on PointNet++\cite{qi2017pointnet++} as the backbone. The network enlarges the grasp-aware region and takes the initial grasp closure local region as an anchor region. Based on this design, it jointly analyzes the geometric features within the initial grasp closure region and the surrounding spatial geometric context, enabling more accurate and robust prediction of grasp feasibility, as well as refinement feasibility.

To validate the effectiveness of the proposed method, we conduct extensive grasping experiments in both simulation and real-world environments across multiple scenarios. The experimental results demonstrate that our method achieves strong stability while maintaining good generalization capability. The task pipeline is shown in Fig.~\ref{fig:task-pipeline}. 

The main contributions of this work are as follows:
\begin{itemize}
    \item We propose a superquadric-based similarity matching method that approximates the overall geometry of a target object from its incomplete point cloud using superquadric coefficients, retrieves multiple globally similar primitives from a database to enhance its geometric representation, and transfers grasp candidates from the matched primitives to the target object.
    
    \item We propose an anchor-guided evaluation-refinement network that builds a grasp-centered contextual representation by integrating the initial closing region with its surrounding spatial neighborhood, enabling more reliable grasp assessment and local refinement.
    
    \item We conduct extensive experiments in both simulation and real-world settings to verify the effectiveness of the proposed method. In particular, all training data are generated in simulation and the model is directly transferred to real-world scenes without any real-world fine-tuning.
\end{itemize}

The code and data associated with this work will be made publicly available.
% Grasping has long been a focus of robotics research. With the rapid development of computer vision and neural networks, grasping methods have become increasingly diverse. 
\section{Related Work}
Grasping has long been a focus of robotics research. With the rapid development of computer vision and neural networks, grasping methods have become increasingly diverse. In this section, we introduce superquadrics and discuss two primary research directions in robotic grasping\cite{review}.
\subsection{Superquadrics}
\begin{figure}[!t]
    \centering
    \includegraphics[width=0.7\columnwidth]{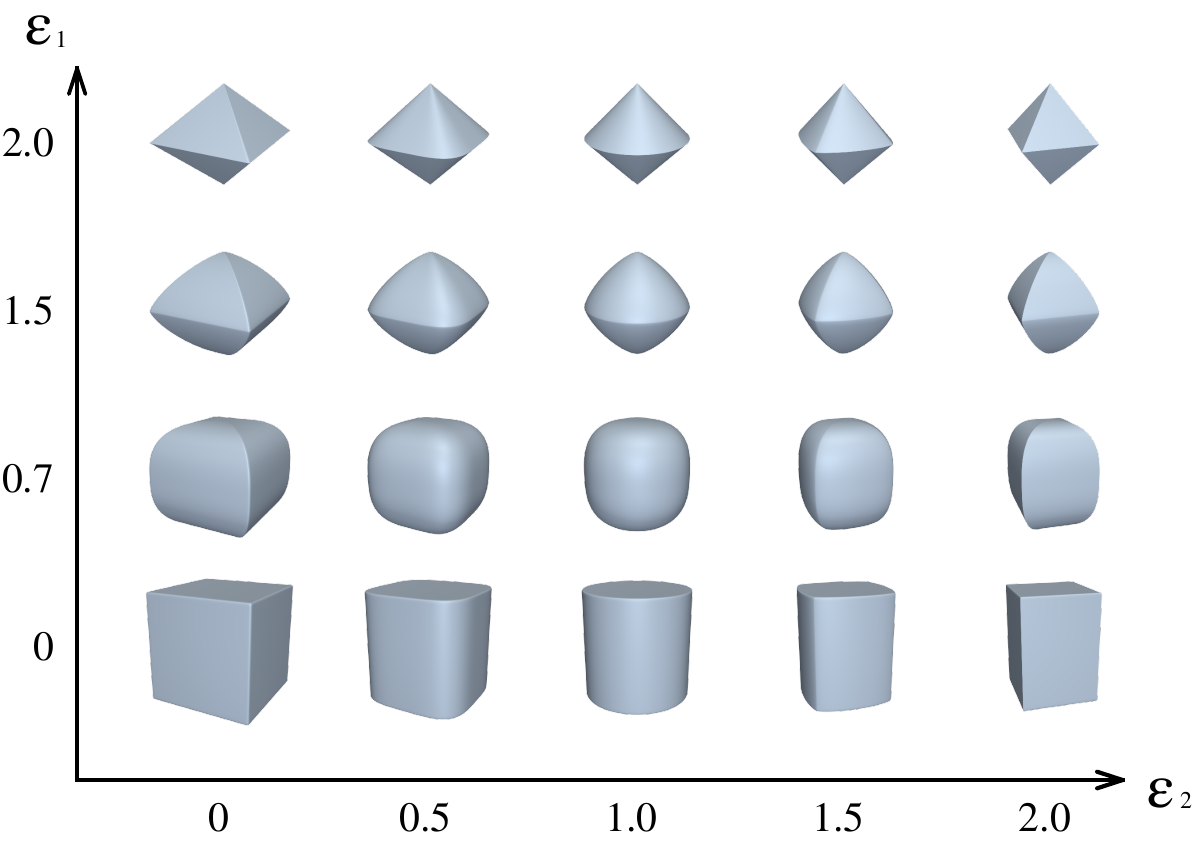}
    \caption{Convex superquadric shapes under different parameter settings.}
    \label{fig:superquadric}
\end{figure}

Superquadrics\cite{ems,barr1981superquadrics,Superquadrics-Revisited} provide a unified representation for regular geometric primitives such as cubes, cylinders, spheres, and ellipsoids using only a small number of parameters. 
A superquadric can be implicitly represented as
\begin{equation}
\left[
\left(
\left|\frac{x}{a_x}\right|^{\frac{2}{\epsilon_2}}
+
\left|\frac{y}{a_y}\right|^{\frac{2}{\epsilon_2}}
\right)^{\frac{\epsilon_2}{\epsilon_1}}
+
\left|\frac{z}{a_z}\right|^{\frac{2}{\epsilon_1}}
\right]
= 1,
\label{eq:superquadric}
\end{equation}

\noindent where $x$, $y$, and $z$ denote the coordinates of the points on the surface in its local coordinate frame. 
The parameters $a_x$, $a_y$, and $a_z$ represent the dimensions of the superquadric along the three principal axes $X$, $Y$, and $Z$, respectively, i.e., the lengths of the three semi-axes. 
In addition, $\epsilon_1$ controls the shape of the longitudinal section along the $z$-axis, while $\epsilon_2$ controls the shape of the cross section in the $xy$-plane. 
When $\epsilon_1, \epsilon_2 \in (0, 2]$, the superquadric surface is convex, as shown in Fig.~\ref{fig:superquadric}.

To fully characterize a superquadric in 3D space, 5 geometric parameters and a 6-DoF pose are required:
\[
\{a_x, a_y, a_z, \epsilon_1, \epsilon_2, \bm{T_{sq}}\},
\]
where \(\bm{T_{sq}} = (\bm{R_{sq}}, \bm{t_{sq}})\) denotes the pose of the superquadric, \({R_{sq}}\) represents its orientation, and \({t_{sq}}\) denotes its translation.
Superquadrics are widely applied in object modeling, shape approximation, and scene understanding due to their low-dimensional representation and strong geometric interpretability. 
They are particularly effective for modeling objects with nearly convex structures, and the recovered parameters can directly convey the graphical, orientational, and shape characteristics of the objects.

\subsection{Two-Stage Grasping}
Two-stage grasping methods typically decompose the problem into grasp candidate generation and grasp evaluation.  GPD\cite{gpd} first samples grasp candidates in 3D space, labels them using force closure analysis \cite{force-closure}, and trains a CNN to predict grasp feasibility from grasp-aligned projections. Method \cite{pointnetgpd} further improves this paradigm by directly encoding the point cloud within the grasp region using a PointNet-based architecture for grasp quality prediction. Later, 3DSGrasp\cite{mohammadi20233dsgrasp} introduced 3D reconstruction to compensate for missing geometric information in point clouds, thus improving both the generation of grasp candidates and the accuracy of grasp feasibility analysis. However, due to the continuous and high-dimensional nature of the grasp space, exhaustive spatial sampling is inherently time-consuming and inefficient. To address this issue, methods\cite{multi,goldfeder2009data} proposed to generate grasp candidates through similarity matching by directly matching the observed partial point cloud to database objects. However, operating directly on incomplete observations makes the overall matching and alignment processes less robust, thereby limiting the reliability of the generated grasp candidates.
\subsection{End-to-End Grasping}
To address sampling inefficiency, methods~\cite{fang2020graspnet-1,wang2021graspnessdiscovery,li2024sim,contact,liu2024efficient-end-to-end,ni2020pointnet++} proposed end-to-end network frameworks that take point clouds as input and directly output grasp candidates with quality scores, relying on large-scale training data to ensure good generalization. Methods~\cite{wei2021gpr,zhao2021regnet} further introduced grasp refinement strategies, first analyzing global point cloud information to obtain coarse grasp poses, then refining them using local geometric details, fully leveraging point cloud spatial information and improving grasp success rates. Method~\cite{6-dof} proposed VAE-based models that encode point clouds into implicit features and decode them to generate 6-DoF grasp poses. However, the aforementioned methods still exhibit limited generalization capabilities when handling variations in scene and object shapes.
\section{Problem Formulation}
We formulate the task as tabletop decluttering with a robotic arm equipped with a parallel-jaw gripper, although the proposed framework can also be applied to target-object grasping. Notably, only a single fixed depth camera is deployed within a \(50\,\mathrm{cm} \times 50\,\mathrm{cm}\) workspace to capture the scene point cloud. The model takes as input the object point cloud together with the global scene point cloud \(X \in \mathbb{R}^{n \times 3}\), and outputs a grasp pose \(g\), defined as:
\[
g=\{p, R, w\},
\]
where $R = [r_1, r_2, r_3] \in \mathbb{R}^{3 \times 3}$ is the rotation matrix representing the gripper orientation, $p=(x,y,z) \in \mathbb{R}^{3 \times 1}$ denotes the gripper position,  and $w \in \mathbb{R}$ denotes the gripper width. Our framework generates an initial set of grasp candidates \(G=\{g_1,g_2,\dots\}\) and then selects the final grasp \(g_{\text{best}}\) through subsequent evaluation and refinement. During execution, if any collision occurs, the grasp attempt is directly regarded as a failure.
\section{Method}
\subsection{Overview}
To improve the efficiency and reliability of grasping, we propose SuperGrasp, a new two-stage framework. First, we introduce a geometric similarity matching strategy to generate grasp candidates (see Sec.~IV-B). Second, we propose an end-to-end evaluation and refinement network, termed E-RNet, to evaluate the stability of the grasp and refine it locally (see Sec.~IV-D).

\subsection{Superquadric Similarity Matching}
\begin{figure*}[!t]
\centering
\includegraphics[width=0.88\textwidth]{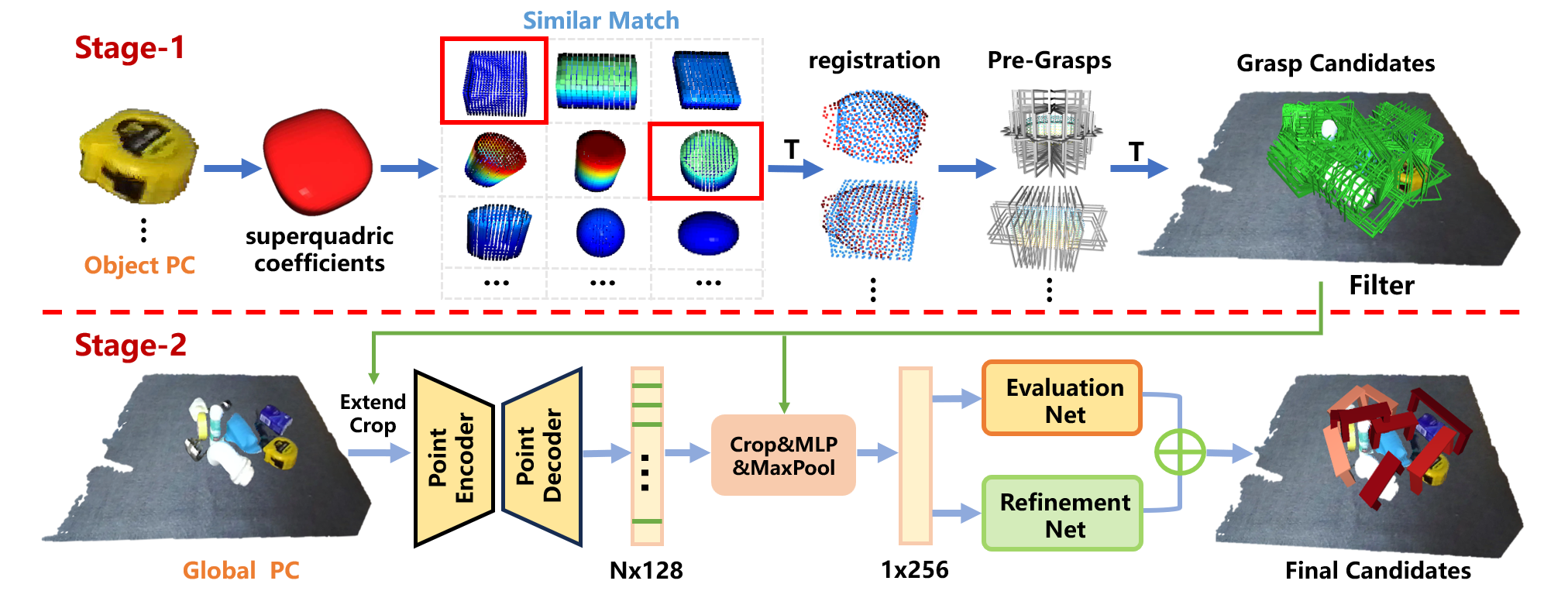}
\caption{Overview: Our framework consists of two stages. In the first stage, we first estimate the superquadric coefficients of the target object and compute their similarity to those of the objects in the database. We then select the top-\(N\) most similar objects and transfer their grasp poses to the target object through a transformation matrix \(T\), followed by a coarse filtering step to remove noisy candidates. In the second stage, we enlarge the grasp-aware region of each initial grasp and feed the cropped regional point cloud into a PointNet++ network for feature extraction. We then further crop the features corresponding to the points within the initial gripper closing region and aggregate them via max pooling to obtain the grasp anchor feature. Finally, the anchor feature is fed into an evaluation network and a refinement network to predict grasp feasibility and refinement feasibility, respectively.}
\label{overview}
\end{figure*}

Superquadrics provide an overall parametric representation of an object using only five shape parameters, $\{a_x, a_y, a_z, \epsilon_1, \epsilon_2\}$, and a 6-DoF pose, $\bm{T}_{sq}$, thereby enabling an alternative grasping paradigm. However, in the single-view setting, point clouds are often incomplete, making it difficult to accurately estimate superquadric coefficients. This leads to biases in both shape and scale, resulting in unstable grasp candidate quality and insufficient coverage. To address this issue, we propose a superquadric-based similarity matching method, as illustrated in Fig.\ref{overview}.

\begin{figure}[!t]
    \centering
    \includegraphics[width=0.85\columnwidth,height=3.8cm]{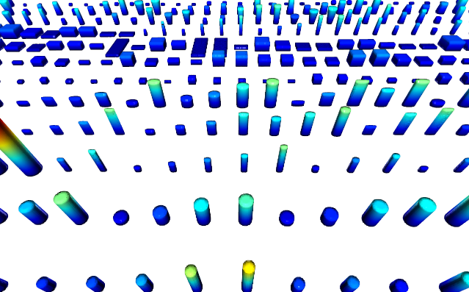}
    \caption{The primitive object database constructed in this work, which contains 1.2k standard geometric objects of various shapes and scales, including cylinders, frustums, elliptical cylinders, and cuboids.}
    \label{fig:dataset}
\end{figure}

\paragraph{Primitive database}
% We first build a database consisting of standard geometric primitives (e.g., cuboids, cylinders, frustums), as shown in Fig.\ref{fig:dataset}. To better fit object scales, we apply proportional scaling to all primitives. For each object, we precompute a set of stable pre-grasp poses $G_1^n=\{g_1,g_2,\ldots\}$ and fix the gripper width $w$, defined as the distance from the inner finger surfaces to the object contact surfaces (set to $0.0125\,\mathrm{m}$ in our experiments). The database contains $1.2\mathrm{k}$ objects in total.
We first build a database of 1.2k standard geometric primitives (e.g., cuboids, cylinders, frustums, and elliptical cylinders), as shown in Fig.~\ref{fig:dataset}. To better cover object scales, we apply proportional scaling to all primitives. For each object, we precompute a set of pre-grasp poses $G_1^n=\{g_1,g_2, \ldots\}$ using family-specific geometric templates that cover the main feasible grasping directions and positions. For each configuration, the corresponding gripper width $w$ is obtained by closing the gripper from an initially open state until the minimum distance between either inner finger surface and the object surface reaches a fixed threshold $\delta=0.0125\,\mathrm{m}$. Depending on the primitive geometry and
  scale, each object contributes 20 to 400 pre-grasp poses, with an average of about 200.
 We further precompute the superquadric coefficients of all database objects using the EMS (Expectation-Maximization Search) fitting method~\cite{ems} and store them offline for subsequent retrieval. 

% \begin{figure}[!t]
%     \centering
%     \begin{minipage}[t]{0.48\columnwidth}
%         \vspace{0pt}
%         \centering
%         \includegraphics[width=\linewidth,height=3.2cm]{superq.png}
        
%         \small (a)
%     \end{minipage}
%     \hfill
%     \begin{minipage}[t]{0.48\columnwidth}
%         \vspace{0pt}
%         \centering
%         \includegraphics[width=\linewidth,height=3.2cm]{dataset.png}
        
%         \small (b)
%     \end{minipage}
%     \caption{Superquadric representation and database construction. (a) Basic geometric objects represented by different superquadric parameters. (b) The primitive object database constructed in this work, which contains 1.5k objects of various shapes and scales, including cylinders, frustums, elliptical cylinders, and cuboids.}
%     \label{fig:two_images}
% \end{figure}
\paragraph{Fast matching}
% To accelerate retrieval, we precompute superquadric coefficients using the EMS (Expectation-Maximization Search) fitting method\cite{ems} for all database objects and store them in a hash table. 
Given a target object, we likewise use EMS to fit its superquadric and obtain the scale parameters $(a_x,a_y,a_z)$ and shape parameters $(\epsilon_1,\epsilon_2)$.
To separately quantify overall size and aspect ratio, we calculate the geometric mean of the scale parameters as follows:
\begin{equation}
m = (a_x a_y a_z)^{1/3},
\end{equation}
and define the normalized scale ratio as:
\begin{equation}
r = \left[\frac{a_x}{m},\frac{a_y}{m},\frac{a_z}{m}\right].
\end{equation}

 The parameterization of superquadrics is non-unique. A surface of a given shape may admit multiple equivalent parameterizations. In our case, this non-uniqueness is detected when $\epsilon_2$ is close to $0$ or $2$ and $a_x$ and $a_y$ are nearly equal. When this condition is satisfied, we generate the corresponding equivalent parametric representation and incorporate both representations into the matching process.

Matching is performed in two steps. 
In the following, the target object is treated as the query, and the superscripts $q$ and $c$ denote the query object and a candidate database object, respectively.
In the first
step, we conduct shape-based coarse filtering using both the superquadric shape parameters and the normalized scale ratio:
\begin{equation}
d_{\epsilon}=
\left\|
[\epsilon_1^{(q)},\epsilon_2^{(q)}]
-
[\epsilon_1^{(c)},\epsilon_2^{(c)}]
\right\|_2,
\end{equation}
\begin{equation}
d_{\mathrm{ratio}}=
\left\|
\log r^{(q)}-\log r^{(c)}
\right\|_2,
\end{equation}
and combine them as:
\begin{equation}
d_{\mathrm{shape}} = w_r d_{\mathrm{ratio}} + w_{\epsilon} d_{\epsilon}.
\end{equation}

All database objects are ranked in ascending order of $d_{\text{shape}}$, and only the top-$K_1$ candidates are retained.

In the second step, we refine the retained candidates using absolute scale differences. 
The overall scale difference is defined as:
\begin{equation}
d_{\mathrm{scale}}=
\left|
\log \frac{m^{(q)}}{m^{(c)}}
\right|,
\end{equation}
the axis-wise scale difference is:
\begin{equation}
d_{\mathrm{abs}}
=
\frac{1}{3}
\sum_{k\in\{x,y,z\}}
\frac{|a_k^{(q)}-a_k^{(c)}|}{\max(a_k^{(q)},a_k^{(c)})}.
\end{equation}
and the final matching score is:
\begin{equation}
s = d_{\mathrm{shape}} + \lambda_s d_{\mathrm{scale}} + \lambda_a d_{\mathrm{abs}}.
\end{equation}

We rank the remaining candidates in ascending order of $s$ and select the top-$K_2$ most similar objects. 
% Their grasp candidates are then transferred to the target object frame using the relative pose $T$ between the fitted candidate and the query superquadrics. 
Their pre-grasp poses are then transferred to the target-object frame using the relative rigid transform ${T = T_{sq}^{q}(T_{sq}^{c})^{-1}}$,
 where $T_{sq}^{q}$ and $T_{sq}^{c}$ denote the fitted poses of the query and candidate superquadrics, respectively.
 In all experiments, we set $w_r=1.0$, $w_{\epsilon}=1.2$, $\lambda_s=1.6$,  $\lambda_a=2.0$, $K_1=50$ and $K_2=5$.

\paragraph{Coarse filtering}
We perform an additional coarse filtering step to remove unreasonable candidates, including: i) those for which the gripper collides with the target object body, ii) those that contain fewer than 50 target points between the gripper fingers, and iii) those for which the angle between the gripper closing direction and the surface normal at the fingertip contact points exceeds $20^\circ$. 

This module yields rich and reliable grasp candidates for the subsequent evaluation and refinement module.
\subsection{Data Collection}
We select 124 graspable objects with various shapes and sizes from YCB\cite{calli2015ycb}, KIT\cite{kasper2012kit}, and DexNet\cite{mahler2016dex} to construct the object dataset. We then build a workspace equipped with a single depth camera in the PyBullet\cite{pybullet} simulation environment. For each scene, five objects are randomly dropped into the workspace, and one of them is randomly designated as the target object. The target object point cloud is obtained from the object mask, and rich grasp candidates are generated using the strategy described in Sec.~IV-B. Finally, one grasp pose is randomly sampled for execution.

\begin{figure}[!t]
    \centering
    \begin{minipage}[t]{0.48\columnwidth}
        \vspace{0pt}
        \centering
        \includegraphics[width=\linewidth,height=3.2cm]{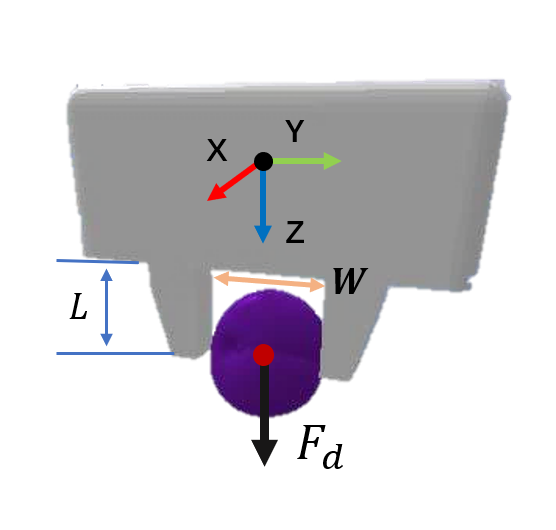}
        
        \small (a)
    \end{minipage}
    \hfill
    \begin{minipage}[t]{0.48\columnwidth}
        \vspace{0pt}
        \centering
        \includegraphics[width=\linewidth,height=3.2cm]{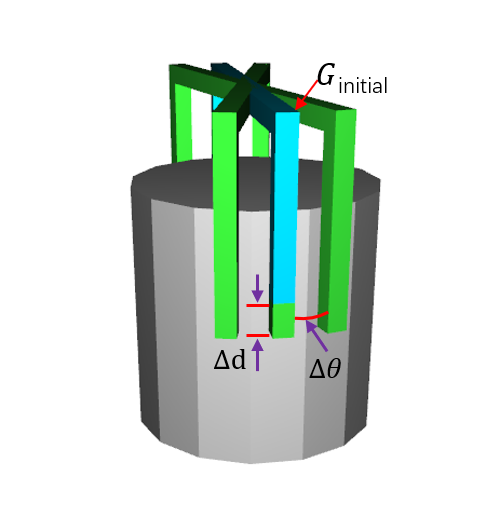}
        
        \small (b)
    \end{minipage}
    \caption{Gripper configuration and refinement strategy. In (a), the $x$-, $y$-, and $z$-axes define the gripper coordinate system, $w$ represents the gripper width, $L$ denotes the fingertip length, and $F_d$ is the applied disturbance force directed vertically downward. In (b), the initial gripper is first extended downward by $0.008\,\mathrm{m}$ along the $z$-axis, and is then rotated by $\pm 15^\circ$ about the $z$-axis.}
    \label{fig:gripper}
\end{figure}

For label assignment, unlike methods\cite{pointnetgpd,gpd}, which grade grasp poses based on analytic metrics, we directly determine the label according to the execution outcome in simulation. Specifically, during execution, if the gripper collides with surrounding objects, the label is directly set to 0. If the gripper successfully lifts the target object and then reorients it to a pose with the gripper approach axis aligned vertically downward, and the grasp remains stable under a downward disturbance force of \(1\,\mathrm{N}\) applied for \(0.5\,\mathrm{s}\), as shown in Fig.~\ref{fig:gripper}, the label is set to 1. All other cases are labeled as 0.

Although the grasp candidates have already been coarsely filtered, slight deviations may still lead to grasp failure in dense cluttered scenes. To exploit the reproducibility of the simulation environment, we further fine-tune the initial grasp pose during data collection. Specifically, we first deepen the grasp by translating it \(0.008\,\mathrm{m}\) along the approach direction, and then additionally rotate it around the approach direction by \([-15^\circ, 0^\circ, 15^\circ]\), resulting in three additional grasp attempts, as shown in Fig. \ref{fig:gripper}. The refinement labels are defined using the same criterion as that of the initial grasp. Following the above strategy, we collect 100K labeled training samples.
\subsection{E-RNet: Evaluation and Refinement Network}
In practical grasp execution, due to the angular deviation

\noindent between the gripper fingers and the local surface normals of the object, a certain degree of relative motion may occur between the gripper and the object during closure. Therefore, it is insufficient to consider only the grasp closing region; the spatial information around this region is also important for reliable grasp evaluation. In addition, although the first stage provides abundant grasp candidates, dense clutter and geometric incompleteness may still cause all initial candidates to fail, in which case slight pose adjustment can improve grasping adaptability. To address these issues, we propose a novel evaluation and refinement network, termed \textbf{E-RNet}. The overall architecture is illustrated in Fig.~\ref{overview}.

To enrich grasp-aware perception, we enlarge the initial grasp closing region. Specifically, the initial grasp pose is first deepened by \(0.008\,\mathrm{m}\) along the approach direction, and then additionally rotated around the approach axis by \(+15^\circ\) and \(-15^\circ\). For stable learning, we transform the global point cloud into the gripper coordinate frame and use the enlarged closing region to crop the region of interest from the global point cloud, yielding a cropped point cloud of size \(N \times 3\). The cropped point cloud is then fed into a PointNet++ backbone to extract point-wise features of size \(N \times 128\).

To model the relationship between the initial grasp region and its surrounding space, we further crop the features corresponding to the original grasp closing region, and use a shared MLP to map the point features into a higher-dimensional feature space before aggregating them by max pooling into a \(1 \times 256\) feature vector. We treat this pooled feature as the \emph{grasp anchor feature}, which encodes not only the geometric characteristics of the local closing region but also its spatial relationship with surrounding points. The grasp anchor feature is then fed into an evaluation classification head implemented as a three-layer multilayer perceptron (MLP) to predict grasp feasibility. The training loss for this branch is defined as the cross-entropy loss:
\begin{equation}
\mathcal{L}_{\mathrm{e}}
=
-\frac{1}{B}\sum_{i=1}^{B}
\left[
(1-\varepsilon)\log p_{i,y_i^{*}}
+
\frac{\varepsilon}{C}\sum_{c=1}^{C}\log p_{i,c}
\right],
\label{eq:lgrasp}
\end{equation}

Unlike methods that fine-tune all grasp candidates in a multi-dimensional pose space to further expand the candidate set\cite{lei2025efficient}, we instead take the initial grasp region as a spatial anchor and explore the grasp feasibility of its neighboring regions, which is more efficient and time-saving. Moreover, unlike Regnet\cite{zhao2021regnet}, which predicts residual corrections with respect to its matched single ground-truth grasp pose, our training strategy allows multiple valid refinement choices, as long as they can lead to a stable grasp. This design provides more refinement flexibility during inference.

Specifically, the grasp anchor feature is simultaneously fed into a refinement head implemented as another three-layer MLP. This branch analyzes the geometric relationships in the surrounding space and predicts the refinement feasibility of the initial grasp pose under three candidate adjustments, namely deepening the grasp by \(0.008\,\mathrm{m}\) along the approach direction and rotating it around the approach axis by \(-15^\circ\), \(0^\circ\) and \(15^\circ\), respectively. This branch is trained using the binary cross-entropy (BCE) loss:
\begin{equation}
\begin{aligned}
\mathcal{L}_{\mathrm{r}}
=& -\frac{1}{3M}
\sum_{i=1}^{M}\sum_{j=1}^{3}
\Big[
y_{ij}\log \sigma(z_{ij}) \\
&\quad + (1-y_{ij})\log \bigl(1-\sigma(z_{ij})\bigr)
\Big]
\label{eq:refine_cls_loss}
\end{aligned}
\end{equation}

During inference, the network predicts the refinement feasibility scores for the three candidate refinement regions in parallel, and the region with the highest score is selected as the final refinement result.

\subsection{Training and Inference Details}
During training, we first add random Gaussian noise to the global point cloud. We then crop the point cloud within the expanded grasp region and downsample it to 960 points. Subsequently, the points within the gripper closing region are further downsampled to 345 points, and their indices in the expanded grasp-region point cloud are recorded for subsequent feature cropping. The network is trained in an end-to-end manner by minimizing the following loss:
\begin{equation}
\mathcal{L} = \lambda_1{\mathcal{L}_e} + \lambda_2{\mathcal{L}_r}.
\end{equation}
where $\lambda_1 = \lambda_2 = 1$. The network is trained on an NVIDIA GeForce RTX 4090 GPU for 100 epochs, with a batch size of 128. We use the Adam\cite{kingma2014adam} optimizer with an initial learning rate of $0.001$ and a weight decay of $3 \times 10^{-4}$. 

During inference, a grasp is executed directly only when its evaluation score exceeds 0.7. If the maximum evaluation score is lower than 0.7 but the maximum refinement score is higher than 0.7, the corresponding initial grasp is further refined before execution.

\section{Experiments}
We conduct extensive comparative experiments in both simulation and real-world environments to demonstrate that: 1) the proposed superquadric similarity matching module can stably generate rich and effective grasp candidates; 2) the proposed evaluation and refinement module can accurately evaluate and refine grasp poses; and 3) the overall framework exhibits good generalization capability and real-time performance. We selected PointNetGPD\cite{pointnetgpd}, Contact-GraspNet \cite{contact}, and Chen et al.\cite{multi} as baselines, where Chen et al.'s method was evaluated only on novel objects since it is designed specifically for novel object settings. Evaluation metrics are as follows:
\begin{itemize}
    \item \textbf{Grasp Success Rate (GSR):} the number of successful grasps divided by the total number of grasp attempts over \(N\) scene trials.
    
    \item \textbf{Task Success Rate (TSR):} the number of successful grasps divided by the total number of objects across \(N\) scene trials.
    
    \item \textbf{Grasp Attempt (GA):} the total number of grasp attempts over \(N\) scene trials, with the number of grasp attempts in each scene limited to at most the total number of objects plus 5.
\end{itemize}
% \textbf{Grasp Success Rate(GSR):} the number of successful grasps divided by the total number of grasp attempts over \(N\) scene trials.

% \textbf{Task Success Rate(TSR):} the number of successful grasps divided by the total number of objects across \(N\) scene trials.

% \textbf{Grasp Attempt(GA):} the total number of grasp attempts over \(N\) scene trials, with the number of grasp attempts in each scene limited to at most the total number of objects plus 5.
    
% \begin{table}[!t]
% \centering
% \caption{Comparison of different methods in simulation}
% \label{tab:sim}

% \footnotesize
% \renewcommand{\arraystretch}{1.15}

% \resizebox{\columnwidth}{!}{%
% \begin{tabular}{lccc|ccc}
% \toprule
% \multirow{2}{*}{Method}
% & \multicolumn{3}{c|}{Grasp success rate}
% & \multicolumn{3}{c}{Task success rate} \\
% \cmidrule(lr){2-4}\cmidrule(lr){5-7}
% & 10 objs & 20 objs & novel & 10 objs & 20 objs & novel \\
% \midrule
% PointNetGPD              & 33.90\% & / & / & / & / & / \\
% Contact-GraspNet         & / & / & / & / & / & / \\
% Ours (S1) + PointNetGPD  & / & / & / & / & / & / \\
% PointNetGPD + Ours (S2)  & / & / & / & / & / & / \\
% Ours w/o RNet            & / & / & / & / & / & / \\
% Ours                     & / & / & / & / & / & / \\
% \bottomrule
% \end{tabular}%
% }
% \end{table}
\begin{table*}[!t]
\centering
\caption{Comparison of different methods in simulation}
\label{tab:sim}
\renewcommand{\arraystretch}{1.12}
\setlength{\tabcolsep}{5pt}
\begin{tabular*}{\textwidth}{@{\extracolsep{\fill}}lccc|ccc|ccc@{}}
\toprule
\multirow{2}{*}{Method} 
& \multicolumn{3}{c}{Seen(obj=10, scene=30)} 
& \multicolumn{3}{c}{Seen(obj=20, scene=30)} 
& \multicolumn{3}{c}{Unseen(obj=10, scene=30)} \\
\cmidrule(lr){2-4} \cmidrule(lr){5-7} \cmidrule(lr){8-10}
& GSR & TSR & GA & GSR & TSR & GA & GSR & TSR & GA \\
\midrule
PointNetGPD\cite{pointnetgpd}            & 78.76\% & 29.67\% & 113 & 59.43\%  & 34.67\%  & 350   & 33.90\% & 6.67\%  & 59  \\
Contact-GraspNet\cite{contact}       & 78.25\% & 80.33\% & 308   & 75.42\% & 82.83\% & 659  & 63.91\% & 69.67\% & 327   \\
Chen et al.\cite{multi}    & -- & -- & -- & -- & -- & -- & 58.14\% & 83.33\% & 430  \\
Ours (S1) + PointNetGPD\cite{pointnetgpd} & 63.34\% & 72.00\% & 341 & 38.42\% & 47.83\%  & 747 & 31.62\% & 37.00\% & 351   \\
PointNetGPD\cite{pointnetgpd} + Ours (S2)& 93.60\% & 53.67\% & 172 & 79.02\%  & 48.33\%  & 367 & 87.85\% & 31.33\% & 107 \\
Ours w/o RNet          & 95.29\% & 87.67\% & 276   & 87.56\% & 84.50\% & 579   & 89.56\% & 88.67\%  & 297  \\
Ours          & 94.02\% & 94.33\% & 301   & 92.82\% & 92.67\% & 599   & 89.03\% & 94.67\% & 319   \\
\bottomrule
\end{tabular*}
\end{table*}
\subsection{Simulation Experiments}
\begin{figure}[!t]
    \centering
    \includegraphics[width=\columnwidth]{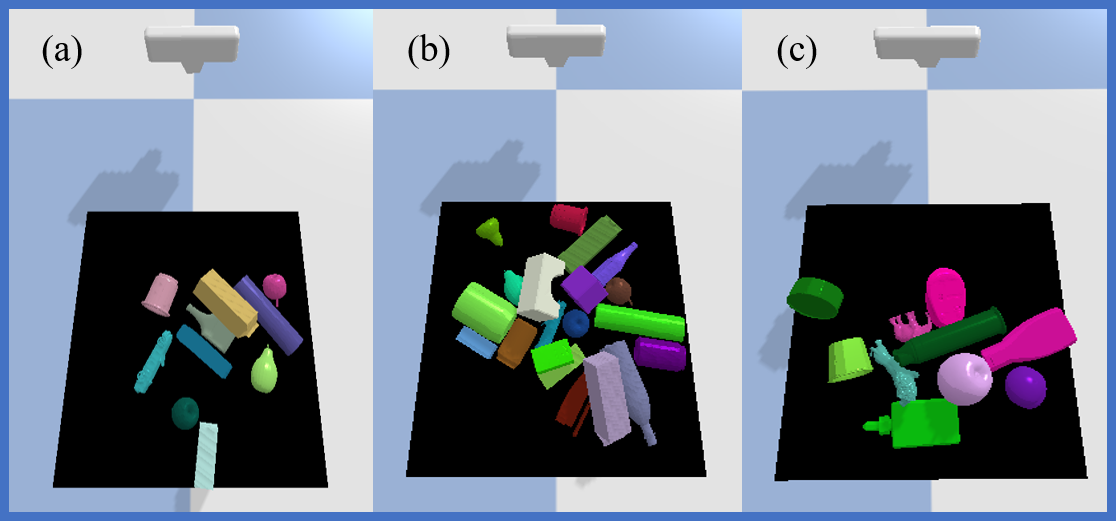}
    \caption{Simulation experiment settings. (a) Scene with 10 seen objects. (b) Scene with 20 seen objects. (c) Scene with 10 novel objects.}
    \label{fig:sim}
\end{figure}

We use the PyBullet\cite{pybullet} simulation environment and deploy a fixed depth camera to capture point clouds in the scene. To thoroughly evaluate the performance of the proposed method, we design three experimental settings: 1) scenes containing 10 objects (moderately cluttered), 2) scenes containing 20 objects (densely cluttered), and 3) scenes containing 10 unseen objects (i.e., objects not included in the training dataset), as illustrated in Fig.~\ref{fig:sim}. For a fair comparison, the same random seeds are used for all methods in each setting to ensure identical scene configurations. Each setting is tested over 30 trials. The average results are reported in Table~\ref{tab:sim}.

As shown in Table~\ref{tab:sim}, in the relatively cluttered scene with 10 objects, our method achieves both an average grasp success rate and an average task success rate of over \(94\%\), demonstrating its strong stability and efficiency. Compared with the structurally similar two-stage method PointNetGPD, our method improves the average grasp success rate by \(15.26\%\) and the average task success rate by \(64.66\%\). Compared with the end-to-end Contact-GraspNet baseline, our method improves the average grasp success rate by \(15.77\%\) and the average task success rate by \(14\%\). These results jointly validate the superiority of the proposed two-stage framework. In the more challenging setting with 20 objects, our method still maintains both the average grasp success rate and the average task success rate above \(92\%\). Specifically, it surpasses PointNetGPD by \(33.39\%\) and Contact-GraspNet by \(17.4\%\) in grasp success rate, while improving the task success rate by \(58\%\) and \(9.84\%\), respectively. These results demonstrate the robustness and generalization capability of our method in highly cluttered environments. When evaluated on unseen objects that are not included in the training set, our method still achieves an average task success rate above \(94\%\). It outperforms PointNetGPD by \(88\%\), Contact-GraspNet by \(25\%\), and Chen et al. by \(11.34\%\) in task success rate, while the average grasp success rate is improved by \(55.13\%\), \(25.12\%\), and \(30.89\%\), respectively. These results further verify the proposed framework's good object-level generalization.
\subsection{Ablation Study}
To more precisely analyze the effectiveness of the proposed framework, we design three ablation settings: 1) \textbf{Ours (Stage-1) + PointNetGPD}, where our grasp candidate generation module is combined with the grasp evaluation module of PointNetGPD to verify the stability and diversity of our sampling strategy; 2) \textbf{PointNetGPD + Ours (Stage-2)}, where the grasp candidate generation module of PointNetGPD is combined with our evaluation and refinement module to validate the accuracy and generalization capability of our second stage; and 3) \textbf{without the refinement branch}, to demonstrate the necessity of the refinement network and quantify its contribution during grasp execution.

We conduct the ablation experiments in the same PyBullet simulation environment under three settings: 1) scenes with 10 objects, 2) scenes with 20 objects, and 3) scenes with 10 unseen objects. Each setting is tested over 30 trials, and the averaged results are reported in Table~\ref{tab:sim}. When our Stage-1 grasp candidate generation module is combined with the PointNetGPD downstream grasp evaluation module, the task success rate is significantly improved compared to PointNetGPD in all settings. Specifically, the average task success rate is improved by \(42.33\%\) in the 10-object setting, \(13.16\%\) in the 20-object setting, and \(30.33\%\) in the unseen-object setting. These results confirm that our sampling strategy can stably provide rich and effective grasp candidates, thereby improving the overall task success rate. However, the average grasp success rate decreases in all settings. We attribute this to the increased number of grasp attempts together with the limited accuracy of the PointNetGPD evaluation module. Notably, when the grasp candidate generation module of PointNetGPD is combined with our Stage-2 module, i.e., grasp evaluation and refinement, the average grasp success rate is substantially improved in all three settings. In particular, in the unseen-object setting, the grasp success rate is improved by \(53.95\%\). Overall, the average task success rate is improved by \(20.77\%\). These results strongly validate the accuracy, robustness, and generalization capability of our evaluation and refinement module. In addition, we observe that the refinement branch has only a limited effect on grasp success rate in the 10-object setting, but brings clear improvements in the more cluttered 20-object setting. Moreover, the average task success rate across the three settings is further improved by \(6.94\%\), which verifies the necessity and effectiveness of the proposed refinement strategy.
\begin{figure*}[!t]
\centering
\includegraphics[width=\textwidth]{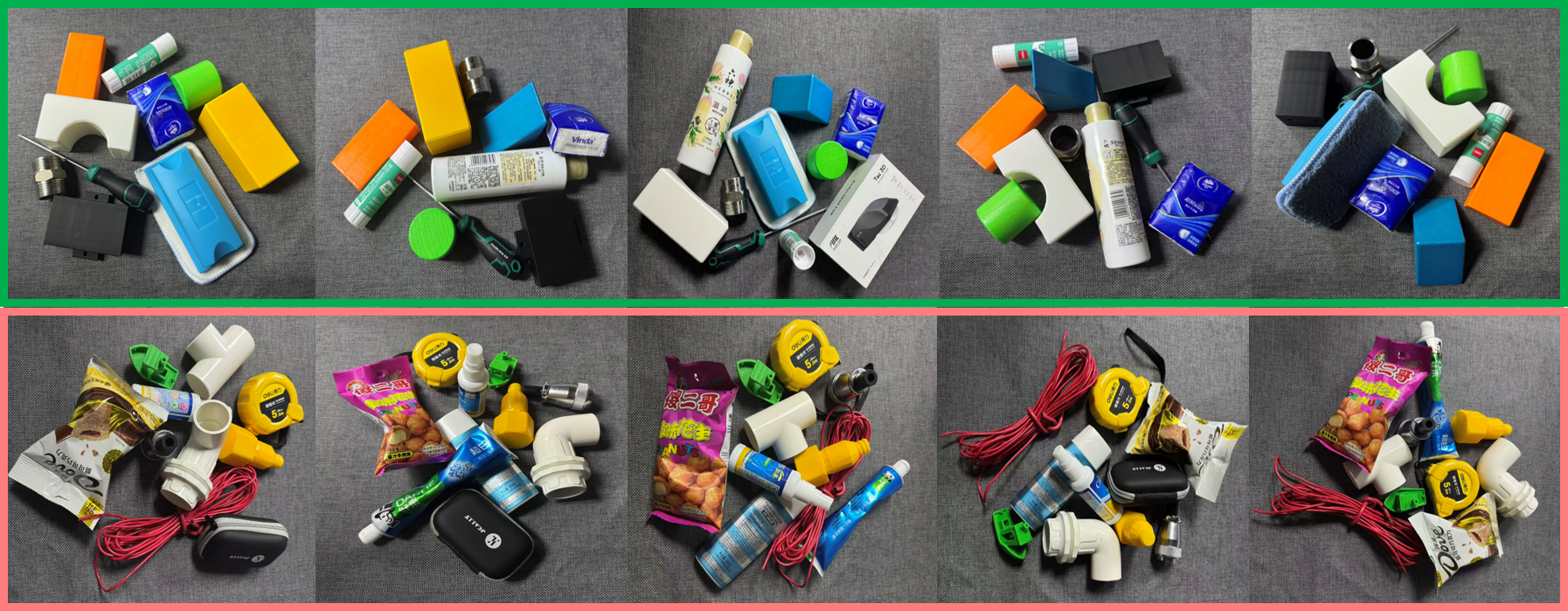}
\caption{Real-world experiment settings. We design two real-world test scenarios: (1) the objects within the green box are similar to those in the training set; and (2) the objects within the red box differ significantly from the training set in both shape and size.}
\label{real}
\end{figure*}
\subsection{Real-World Experiments}
To further validate the generalizability of the proposed method, we construct a real-world grasping test set consisting of 30 \emph{similar} objects and 25 \emph{novel} objects whose shapes and sizes are substantially different from those in the training set. We design two cluttered real-world scenarios for evaluation: 1) randomly placing 10 similar objects and 2) randomly placing 10 novel objects. Each scenario is tested over 10 trials, and some representative scenarios are shown in Fig.~\ref{real}. Specifically, we use an xMate ER7Pro robotic arm, an AG-105-145 gripper, and a ZED~2i camera. The entire system is integrated through ROS, and inverse kinematics is solved using MoveIt!\cite{chitta2012moveit}. All methods employ FoundationStereo\cite{wen2025foundationstereo} to estimate depth from stereo RGB images and then use SAM\cite{kirillov2023segment} to segment the target object. For a fair comparison, all methods are evaluated under the same experimental setup and perception pipeline. The experimental results are reported in Table~\ref{tab:real}.

\begin{table}[!t]
\centering
\caption{Comparison of different methods in the real world}
\label{tab:real}

\footnotesize
\renewcommand{\arraystretch}{1.15}

\resizebox{\columnwidth}{!}{%
\begin{tabular}{lcc|cc}
\toprule
\multirow{2}{*}{Method}
& \multicolumn{2}{c|}{Similar (obj=10, scene=10)}
& \multicolumn{2}{c}{Novel (obj=10, scene=10)} \\
\cmidrule(lr){2-3}\cmidrule(lr){4-5}
& GSR & TSR & GSR & TSR \\
\midrule
PointNetGPD\cite{pointnetgpd}      & 58.82\% & 20.00\% & 52.38\% & 11.00\% \\
Contact-GraspNet\cite{contact} & 78.95\% & 60.00\% & 74.65\% & 53.00\% \\
Chen et al.\cite{multi} & -- & -- & 69.23\% & 90.00\% \\
Ours             & 98.00\% & 98.00\% & 94.12\% & 96.00\% \\
\bottomrule
\end{tabular}%
}
\vspace{-2mm}
\end{table}

\begin{table}[!t]
\centering
\caption{Time cost of the grasping process}
\label{tab:time_cost}

\footnotesize
\renewcommand{\arraystretch}{1.15}

\begin{tabular}{lccc}
\toprule
Method & Stage-1 (s) & Stage-2 (s) & Total (s) \\
\midrule
PointNetGPD\cite{pointnetgpd}      & 9.276 & 0.732 & 10.008 \\
Contact-GraspNet\cite{contact} & --    & --    & 0.333  \\
Chen et al.\cite{multi} & 0.509    & 1.484    & 1.993 \\
Ours             & 0.346 & 0.367 & 0.713  \\
\bottomrule
\end{tabular}
\end{table}
As shown in Table~II, in the \emph{similar-object} setting, our method achieves both a grasp success rate and a task success rate of \(98\%\). Specifically, the average grasp success rate exceeds that of PointNetGPD by \(39.18\%\) and that of Contact-GraspNet by \(19.05\%\), while the average task success rate is higher than that of PointNetGPD by \(78\%\) and that of Contact-GraspNet by \(38\%\). In the novel-object setting, our method still maintains a solid performance, with both the grasp success rate and the task success rate remaining above \(94\%\). In this setting, the average grasp success rate of our method is higher than that of PointNetGPD, Contact-GraspNet, and Chen et al. by \(41.74\%\), \(19.47\%\), and \(24.89\%\), respectively, while the average task success rate is higher by \(85.00\%\), \(43.00\%\), and \(9.00\%\), respectively. These results demonstrate the good generalization capability of the proposed method. 

During the experiments, we also record the average time cost of each stage, as reported in Table~\ref{tab:time_cost}. Compared with the baselines, our method achieves the best real-time performance among all two-stage methods, with runtime only slightly higher than that of the end-to-end network Contact-GraspNet. In addition, we observe that the primary cause of grasp failure in real-world experiments is self-occlusion of the target object or occlusion caused by surrounding objects, which introduces unknown space in the scene and eventually causes collisions between the gripper and the object during execution. Furthermore, objects with severely non-convex geometries can also degrade the quality of grasp sampling, thereby increasing the likelihood of grasp failure.
\section{Conclusion}
We proposed SuperGrasp, a new two-stage framework for single-view object grasping under incomplete geometric observations. It combines superquadric-based similarity matching for effective and diverse grasp candidate generation with an anchor-guided evaluation-refinement network for accurate and robust grasp evaluation and local refinement. Extensive experiments demonstrate the effectiveness of both stages and show that this combination makes single-view grasping more stable and efficient, while enabling solid performance and good generalization across diverse scenarios. Notably, all training data are collected entirely in simulation, and the model transfers directly to real-world scenes without any real-world fine-tuning.
% \bibliographystyle{IEEEtran}
\bibliographystyle{ieeetr}   
\bibliography{References_v2} 
% \end{thebibliography}

\newpage
\vfill

\end{document}